\documentclass{article}

\PassOptionsToPackage{numbers, compress}{natbib}

     \usepackage[preprint]{neurips_2019}

\usepackage{microtype}
\usepackage{graphicx}
\usepackage{subfigure}
\usepackage{booktabs} 
\usepackage{amsmath}
\usepackage{amsthm}	
\usepackage{amsfonts}

\usepackage[utf8]{inputenc} 
\usepackage[T1]{fontenc}    
\usepackage{hyperref}       
\usepackage{url}            
\usepackage{booktabs}       
\usepackage{amsfonts}       
\usepackage{nicefrac}       
\usepackage{microtype}      

\usepackage{amssymb}

\usepackage{algorithm}
\usepackage{algorithmic}
\usepackage{threeparttable}

\title{Spurious Local Minima Are Common for Deep Neural Networks with Piecewise Linear Activations}

\author{%
  Bo Liu\\
  College of Computer Science, Faculty of Information Technology\\
  Beijing University of Technology, Beijing, China \\
  \texttt{liubo@bjut.edu.cn} \\
}

\begin{document}

\maketitle

\begin{abstract}
 In this paper, it is shown theoretically that spurious local minima are common for deep fully-connected networks and convolutional neural networks (CNNs) with piecewise linear activation functions and datasets that cannot be fitted by linear models. A motivating example is given to explain the reason for the existence of spurious local minima: each output neuron of deep fully-connected networks and CNNs with piecewise linear activations produces a continuous  piecewise linear (CPWL) output, and different pieces of CPWL output can fit disjoint groups of data samples when minimizing the empirical risk. Fitting data samples with different CPWL functions usually results in different levels of empirical risk, leading to prevalence of spurious local minima. This result is proved in general settings with any continuous loss function. The main proof technique is to represent a CPWL function as a maximization over minimization of linear pieces. Deep ReLU networks are then constructed to produce these linear pieces and implement maximization and minimization operations. 
\end{abstract}

\section{Introduction} \label{Introduction}
Although deep neural networks, such as deep fully-connected feedforward networks and CNNs (e.g. \cite{AlexNet,VGG}), have been successfully applied to many disciplines including computer vision and natural language processing, the non-convex loss landscape of deep neural networks, especially whether sub-optimal (spurious) local minima commonly exist, is still not well understood in theory. Since gradient-based local search methods (such as gradient descent) can easily get stuck in spurious local minima, understanding the existence of spurious local minima is very important for deep network training and has drawn the attention of many researchers (see \cite{GlobalLandscapeOverview} and section \ref{relatedwork} for overviews).

Most deep neural networks popular today use the ReLU activation function.  However, the theoretical understanding of spurious local minima for ReLU networks is still limited. 
For one-hidden-layer ReLU networks, the statement that spurious local minima exist has been shown in the studies of \cite{SafranShamir18,Swirszcz,ChulheeYun19,CRITICALPOINTS,BoundsDescentPathsICLR20}.  These works either constructed concrete data examples and networks or performed experiments to demonstrate the existence of spurious local minima.
Recently, \cite{Dacheng_Tao2020,Ding,Goldblum2020} exhibited existence of spurious local minima for deep fully-connected  ReLU networks. They constructed deep ReLU networks that locally emulate linear neural networks to fit all data samples and thus inherit the local minima of linear neural networks. All ReLU neurons are active at the constructed spurious local minima such that  networks behave linearly. Therefore, the constructed  spurious local minima are located in unimportant regions in parameter space and usually not encountered in practice when optimizing with appropriate initializations. The following theoretical questions remain unanswered or require a deeper understanding before this work.  Why do spurious local minima exist?  Are they common and widespread in parameter space and not restricted to those specific locations in existing constructions? Do spurious local minima exist for any practical dataset and network?  The goal of this work is to understand  why spurious local minima arise  and construct spurious local minima that are commonly seen for deep fully-connected ReLU networks and CNNs.

In this paper, it is shown that spurious local minima are common in deep fully-connected networks and CNNs with piecewise linear activations.  Starting with observations drawn from a motivating example, the essence of spurious local minima and why they are common in deep fully-connected networks and CNNs with piecewise linear activations are explained.
The key insight is that each output neuron of deep neural networks with piecewise linear activation functions produces a CPWL output. When minimizing the empirical risk, each linear piece of the CPWL function can optimally fit different disjoint group of data samples, and different fitting patterns usually result in different levels of empirical risk, giving rise to the prevalence of spurious local minima.
Explicit constructions of deep fully-connected ReLU networks and CNNs are then presented to show the existence of spurious local minima in general settings with any continuous loss function and any dataset that cannot be fitted by linear models. The main idea is to represent a locally optimal CPWL predictor as a maximization over minimization of constituent linear pieces, and construct ReLU networks to produce linear pieces and implement maximization and minimization operations. 
To the author's best knowledge, this is the first theoretical work showing the prevalence of spurious local minima in deep fully-connected ReLU networks and CNNs, which deepens our understanding of their loss landscape.

This paper is organized as follows. Section \ref{relatedwork} is related work. Section \ref{section2} describes the deep ReLU networks and convolutional neural networks, and gives some mathematical preliminaries on spurious local minima. In section \ref{section4}, a motivating example is provided to illustrate the essence of spurious local minima and why they are common. Section \ref{section5} presents the main results and their proof sketches. In section \ref{section6}, the detailed proofs of theorems are given. Finally, conclusions and future directions are provided.

\section{Related Work}\label{relatedwork}

It has been shown that there are no spurious local minima for some networks and learning models, including deep linear networks \cite{Baldi,Kawaguchi_nips16,HaihaoLu,Laurent18,GLOBALOPTIMALITY,Nouiehed,Zhang}), matrix completion and tensor decomposition (e.g., \cite{Matrixcompletion},  one-hidden-layer networks with quadratic activations \cite{Soltanolkotabi,QuadraticActivation}, deep linear residual networks \cite{HardtMa}, deep quadratic networks \cite{DeepQuadratic} etc.

For one-hidden-layer ReLU networks, by constructing specific examples of networks and(or) data samples, the existence of spurious local minima has been demonstrated in the studies of  \cite{SafranShamir18,Swirszcz,CRITICALPOINTS,ChulheeYun19,Ding,BoundsDescentPathsICLR20} etc.
Instead of specific constructions, the following works discussed local minima for one-hidden-layer ReLU networks by analyzing the conditions for their existence. \cite{SoudryCarmon} gave the conditions under which a differentiable local minimum has zero loss and thus is global. \cite{Multilinear} showed that ReLU networks with hinge loss can only have non-differentiable local minima and gave the conditions for their existence for linear separable data. For over-parameterized one-hidden-layer ReLU networks,  \cite{SafranShamir16} showed that there is a high probability of initializing in a low-loss basin. \cite{SoudryHoffer} exhibited that, given standard Gaussian data and Gaussian weight initialization, the volume of differentiable regions containing spurious local minima is exponentially vanishing.
\cite{globallosslandscape_part1,Dacheng_Tao2020} proved that local minima are global in each differentiable region,  and \cite{globallosslandscape_part1} gave the necessary and sufficient conditions for the existence of spurious local minima in one-hidden-layer ReLU networks with MSE (mean square error) loss.
\cite{BoundsDescentPathsICLR20} constructed a complex data pattern to produce cup-shaped non-differentiable local minima for one-hidden-layer ReLU networks.

For deep neural networks,  \cite{Dacheng_Tao2020,Ding,Goldblum2020} showed that spurious local minima exist for fully-connected deep neural networks with arbitrary piecewise linear activations and a general class of differentiable loss functions. These works constructed examples of spurious local minima in which all ReLU neurons are active such that the deep neural networks are reduced to linear predictors, and therefore these constructions are relatively trivial. 

Besides exploring the existence of spurious local minima in the empirical risk of neural networks, the loss landscape of neural networks has been studied from other perspectives. 
The non-existence of spurious valley (bad basins) for ultra-wide networks was explored in \cite{Venturi,Nguyen19,Li,Nguyenvalleys,Ding}.
Some researches tried to reshape the loss landscape by modifying the network architecture or using regularization. 
\cite{LossSurfaceBinary,AddingOneNeuron,Kawaguchi19} showed that by adding a single-layer network or even a single special neuron in the shortcut connection, every local minimum becomes global. \cite{landdesign,Gao,Feizi} designed new losses or special networks to make local minima global. \cite{Deforming-Loss-Surface} deformed loss surfaces to enhance the gradient descent optimizer’s ability to filter out sharp minima. \cite{ResNetsProvablyBetter,KawaguchiBengio} proved that losses of all local minima in ResNets are no worse than those of global minima in corresponding shallow linear predictors. \cite{piecewiseconvex} showed that the loss surface of a feed-forward ReLU network regularized with weight decay is strongly convex in a special region. 
\cite{Choromanska15} used spin glass models in statistical physics to analyze the loss landscape, which greatly ignore the nonlinear nature of deep neural networks.
\cite{Spectrum,RandomMatrix} used random matrix theory to study the spectrum of Hessian matrix of loss. \cite{Mei,Zhou_Feng} studied the landscape of expected loss. 
\cite{Bianchini} showed the topological expressiveness capability of deep networks.

Some works  studied  loss landscape of deep neural networks empirically.
\cite{attackingthesaddle} performed various experiments to show that training can get stuck in plateaus and advocated that saddle points are the main source of difficulty for optimizing deep networks. \cite{Topologygeometry} discovered that the loss landscape becomes bumpy for low empirical risks, indicating that spurious local minima are common in low-loss regions. \cite{Goodfellow,Poggio,Visualizing,largscalelandscape} explored the visualization of loss landscape.  
By visualizing random slices of loss surface, \cite{Visualizing} observed that the loss surface of deep feedforward networks transitions from nearly convex to chaotic with increasing depth. 
\cite{RINGINGRELUS2021} explained this observation by spectral analysis.
\cite{EssentiallyNoBarriers,modeconnectivity,Low-loss-connection} revealed the so-called mode connectivity phenomenon in which different global minima are connected by low-loss curves for over-parameterized networks.

Understanding the landscape of empirical loss focuses on the geometry side of neural network optimization. Another line of research studies optimization of neural networks by exploring the convergence of gradient descent. For example,  \cite{Yuandong_Tian_17,Zhong17,Yuanzhi_Li_17,Soltanolkotabi,Zhou_Liang_19,OnehiddenlayerReLUviaGD} studied the convergence of gradient descent for one-hidden-layer ReLU networks. Some recent works (e.g., \cite{SimonDu19,Allen-Zhu,QuanquanGu}) showed that gradient descent locally converges for deep networks if they are sufficiently wide, the step-size is small and the weights are initialized in a small region. 
Instead of local analysis in small regions for over-parameterized networks, this paper is more general by considering existence of spurious local minima at any possible locations in parameter space for ReLU networks of regular size.  In a related recent line of work, it has been shown that the training of over-parameterized deep neural networks can be characterized by the training dynamics of kernel regression with the neural tangent kernel (NTK) \cite{NTK}.

Neural networks with piecewise linear activations output continuous piecewise linear functions of inputs. 
The number of linear regions in such piecewise linear outputs was discussed in (e.g. \cite{Montufar14,BoundingCounting,Hanin19,Hanin19nips}) for deep fully-connected neural networks and in \cite{linearpiecesCNN} for CNNs.

The works most related to this paper are \cite{Dacheng_Tao2020,Ding,ChulheeYun19,Goldblum2020}. The result of  \cite{ChulheeYun19} is similar to \cite{Dacheng_Tao2020} and only  considers one-hidden-layer networks.  \cite{Dacheng_Tao2020},  \cite{Ding} and \cite{Goldblum2020} dealt with deep networks with  piecewise linear activation functions, and \cite{Dacheng_Tao2020} is more general and complete.  The main differences between this paper and these works include:
1. \cite{Dacheng_Tao2020,Ding,ChulheeYun19,Goldblum2020} use a single linear predictor to fit all data samples, resulting in a single spurious local minimum. In comparison, this work uses continuous piecewise linear predictor to fit disjoint group of samples. Thus, the predictors in this paper are more flexible and can have many different fitting patterns, leading to prevalence of spurious local minima. 
2. In \cite{Dacheng_Tao2020,Ding,ChulheeYun19,Goldblum2020} all ReLU neurons in the constructed networks are active (thus the networks are reduced to linear predictors), which is unusual for practical ReLU networks and the constructed spurious local minima are rarely encountered in practical optimization. In comparison, some ReLU neurons in the constructions of this paper can be inactive and the resulted spurious local minima are more common in parameter space. 
3. The techniques for proving that the constructed  local minima are spurious are different. In \cite{Dacheng_Tao2020,ChulheeYun19}, it is shown that there exist perturbation directions which cause the empirical risk to decrease. In \cite{Goldblum2020}, the issue of whether the constructed local minima are really spurious is merely addressed by providing numerical evidence, and in \cite{Ding} the authors resort to the assumption of realizability to resolve this problem. While in this work, the local minima with lower empirical risks are constructed explicitly.  
4. This work also deals with CNNs, which are not considered in \cite{Dacheng_Tao2020,Ding,ChulheeYun19,Goldblum2020}.

\section{Notations and Preliminaries} \label{section2}

\subsection{Notations} 
$ M_{i,.} $ and $ M (i,.) $ denote the $ i $th row of matirx $ M $, $ M_{.,i} $ and $ M (.,i) $ denote the $ i $th column of $ M $. $M_{i, j}$ and $M_{ij}$ represent the $(i, j)$-th entry of $M$.  $v_{i}$ denotes the $i$th component of vector $\textbf{v}$. $ [N] $ is the abbreviation of  $ \left \{ 1, 2, \cdots, N \right \} $.

\subsection{Deep ReLU Networks} \label{section2.1}
Deep fully-connected feedforward ReLU networks are described in this subsection. 
Consider a training set comprised of samples
$ \left\{\left(\mathbf{x}_{1}, \mathbf{y}_{1}\right),\left(\mathbf{x}_{2}, \mathbf{y}_{2}\right), \ldots,\left(\mathbf{x}_{N}, \mathbf{y}_{N}\right)\right\} $, where $ \mathbf{x}_{i} \in \mathbb{R}^{d_x}, \mathbf{y}_{i} \in \mathbb{R}^{d_{y}} $ are respectively the  input and output of the $ i $th sample. 
Let $ L $ be the number of layers in a network. Denote the weight matrix and bias of the $ j $th layer as $ W^{j} $ and $ \mathbf{b}^{j} $ respectively with $ W^{j} \in \mathbb{R}^{d_{j} \times d_{j-1}}, \ \mathbf{b}^{j} \in \mathbb{R}^{d_{j}} $, where $ d_{j} $ is the number of neurons (or width) in the $ j $th layer. The output of the $ i $th data sample at the $ l $th layer, denoted as $ \mathbf{y}^{l, i}$, is computed by 
\begin{equation} \label{eq_p1}
	\mathbf{y}^{l, i} = \sigma\left( W^{l} \mathbf{y}^{l-1, i}+\mathbf{b}^{l} \right) , \ \ ( l=1,2, \cdots, L-1)  ,
\end{equation} 
with $ \mathbf{y}^{0, i} :=\mathbf{x}_{i}   $. $ \sigma $ is the activation function which computes component-wise, and for ReLU networks $ \sigma(x)=max(0, x) $.
The output vector of the network is 
\begin{equation*}
	\hat{\mathbf{y}}^{i}  = W^{L} \mathbf{y}^{L-1, i} +\mathbf{b}^{L}.
\end{equation*}

To represent the activation states of  ReLU neurons, let us introduce a diagonal matrix $ I^{l,i} \in \mathbb{R}^{d_{l} \times d_{l}} \ \ (1 \leq l \leq L-1) $ for each sample and each ReLU layer whose diagonal entries are defined as $ I_{k, k}^{l, i}=1 $ if $ W^{l} (k,:) \mathbf{y}^{l-1, i} + b^{l}_k  > 0 $ and $ I_{k, k}^{L, i}=0 $ otherwise. Consequently, there is $ \mathbf{y}^{l, i}=I^{l, i} (W^{l} \mathbf{y}^{l-1, i}+\mathbf{b}^{l}) \ \ (1 \leq l \leq L-1) $.
Network output can then be written as
\begin{equation} \label{eq_p2}
	\hat{\mathbf{y}}^{i}  =W^{L} I^{L-1, i} \left( W^{L-1}  \cdots \left( W^{1} \mathbf{x}_{i} + \mathbf{b}^{1} \right)
	+ \cdots  + \mathbf{b}^{L-1}\right) + \mathbf{b}^{L} .
\end{equation}

The empirical risk for network training is 
\begin{equation} \label{eq_p3}
	R (W^{1}, \mathbf{b}^{1}, \cdots, W^{L}, \mathbf{b}^{L} )=\frac{1}{N} \sum_{i=1}^{N} l \left(\hat{\mathbf{y}}^{i}, \mathbf{y}_{i}\right) ,
\end{equation}
where $ l $ is the loss function, such as the MSE loss and cross-entropy loss.

To be succinct, the biases will be absorbed in weights (correspondingly, the output vector of each layer will be augmented with a scalar 1). The output of the $ l $th layer is then written as $ \mathbf{y}^{l, i}=\sigma\left( W^{l} \mathbf{y}^{l-1, i}\right) , \ 1 \leq l \leq L-1 $, and the empirical risk is abbreviated as $ R(W)=R (W^{1},\cdots, W^{L} ) $.

\subsection{Convolutional Neural Networks} \label{section2.2}

A CNN usually includes some convolutional layers, pooling (max pooling or average pooling) layers and fully-connected layers. Fully-connected layers are defined in the same way as those for deep fully-connected networks.

Convolutional layers are designed to take advantage of the translational invariance inherent in such  data  as images. Suppose the $ l $th layer is a convolutional layer, the output of each neuron in layer $ l $ is obtained by first taking the inner product between a filter of layer $ l $ and a patch of neighboring neurons at layer $ (l-1) $, adding the bias and then applying the activation function. 
Let $ P_l $ and $ s_l $ be respectively the number of patches and the size of each patch at layer $ l $. For input $ \mathbf{x} $, denote $ \left \{ \mathbf{y}^l_1, \cdots, \mathbf{y}^l_{P_l} \right \} \in \mathbb{R}^{s_l}  $ the set of patches at layer $ l $. Denote by $ T_l $ the number of convolutional filters and by $ W^l = [\mathbf{w}^l_1, \cdots, \mathbf{w}^l_{T_l} ]^T \in \mathbb{R}^{T_l \times s_{l-1}}  $  the corresponding weight matrix of layer $ l $. Each row of $ W^l $ corresponds to one filter. The number of neurons at layer $ l $ is $ n_l = T_l P_{l-1} $. Given the $ p $th $ (p \in [P_{l-1}]) $ patch $ \mathbf{y}^{l-1}_p $ and the $ t $th $(t \in [T_l]) $ filter $ \mathbf{w}^l_t $, the output of corresponding neuron in layer $ l $ is as follows,
\begin{equation} \label{eq2_2_1}
	y^l_h = \sigma ( \left \langle  \mathbf{w}^l_t, \mathbf{y}^{l-1}_p  \right \rangle + b^l_h ) ,
\end{equation}
where $ h = (t-1)P_{l-1} + p $, $ \left \langle  \right \rangle $ denotes inner product. 
By absorbing the biases into weights and using the notation $ * $ to denote convolution, the output of the $ l $th layer is abbreviated as $ \mathbf{y}^l = \sigma (W^l * \mathbf{y}^{l-1} ) $.

A layer $ l $ is called a max-pooling layer if the output of its $ p $th neuron $ y^l_p $ is computed by
\begin{equation} \label{eq2_2_2}
	y^l_p = \text{max} ( (y^{l-1}_p)_1, \cdots, (y^{l-1}_p)_{s_{l-1}} )  , \ p \in [P_{l-1}],
\end{equation}
where $  (y^{l-1}_p)_i $ is the $ i $th element in the $ p $th patch of layer $ (l-1) $.  Similarly, if the average of every patch from the previous layer is computed,
\begin{equation} \label{eq2_2_3}
	y^l_p = \text{mean} ( (y^{l-1}_p)_1, \cdots, (y^{l-1}_p)_{s_{l-1}} )  , \ p \in [P_{l-1}],
\end{equation}
then layer $ l $ is called a average-pooling layer.

Given the CNN output  $ \hat{\mathbf{y}}^{i} $ for a sample $ \mathbf{x}_{i} $, the empirical risk for network training is still given by
\eqref{eq_p3}.

\subsection{Definitions: Local Minima and Spurious Local Minima}\label{section2.3}

$  \hat{W} =  (\hat{W^1}, \cdots, \hat{W^L})  $ is called a global minimum of loss
$ R(W) $ if and only if $ \forall W , \ R(\hat{W}) \leq R(W) $. $ \bar{W} $ is called a critical (or stationary) point of loss $ R(W) $ if and only if  $ \frac{\partial R}{\partial W} (\bar{W}) =0 $. $ W^\ast $ is defined as a local minimum of loss $ R(W) $ if and only if there exists an open set $ D $ which contains $ W^\ast $ such that $ \forall W \in D, \ R(W^\ast) \leq R(W) $, and $ W^\ast $ is called a strict local minimum if for any other $  W \in D, \ R(W^\ast) < R(W) $. A spurious local minimum is a local minimum $ W^\ast $ satifying $ R(\hat{W}) < R(W^\ast) $. The local maximum can be defined in a similar way. A saddle point is defined as a critical point which is neither a local minimum nor a local maximum. 

One main goal of this paper is to construct spurious local minima that are frequently encountered for the empirical risks of deep fully-connected networks and CNNs given in \eqref{eq_p3}.

\section{A Motivating Example}\label{section4}

In this section, an intuitive example will be presented to illustrate why spurious local minima are common in deep ReLU networks, and subsequently the main ideas of this paper will be drawn. The key insight is that for neural networks with piecewise linear activation functions, each output neuron produces a CPWL function of input that is optimized to fit data samples by minimizing the empirical risk. The ReLU activation function is piecewise linear with two pieces. When taking CPWL functions as input, the outputs of both affine transformations and ReLU activations are still CPWL. Therefore, any neuron in the hidden and output layers of ReLU networks outputs a CPWL function. 

Consider the regression of data samples lying on a curve with ReLU networks. In fig. 1, there are $ N=40 $ samples evenly spaced in the interval [-1 1] on $ x $ axis, and the target for each sample is generated by $ y=x^2 $. The empirical risk is
\begin{equation*}
	R=\frac{1}{N} \sum_{i=1}^{N} l(f(x_i), y_i),
\end{equation*}
where $ f(x_i) $ denotes the output of a ReLU network that takes a sample $ x_i $ as input. The loss function $ l $ is assumed to be MSE loss.  

Let us partition the samples into several disjoint groups. Therefore,
\begin{equation*}
	R(f)=\frac{1}{N} \sum_{j=1}^{P}\sum_{i=1}^{N_j} l(f(x_i), y_i),
\end{equation*}
where $ P $ is the number of groups and $ N_j $ is the number of samples in the $ j $th group. Fig. 1(a) shows a case of partitioning into two groups. Then, optimal line segments $ f^\ast_j \ (j\in [P]) $ are used to fit different group of samples by minimizing the MSE loss,
\begin{equation}\label{eq4_0}
	f^\ast_j = \text{arg min}_{f_j} \sum_{i=1}^{N_j} l(f_j(x_i), y_i) , \ j\in [P] .
\end{equation}
Such optimal line segments can be obtained by the least square method. 

Excluding the trivial dataset where all samples are identical,  between any two neighboring disjoint groups of samples there is a middle region on the $ x $ axis in which there have no samples. If neighboring optimal line segments intersect at points lying within such middle regions, then as shown in fig. 1(a), a CPWL function $ f $ is formed in which all samples of the $ j $th group are predicted by the $ j $th optimal line segment and not affected by others. If intersection points of neighboring optimal line segments are not located in corresponding middle regions, one can use a auxiliary line segment within the middle region to connect neighboring optimal line segments, as shown in fig. 1(b). Therefore, CPWL functions can always be constructed to fit data samples. Such CPWL functions can  be generated by ReLU networks with appropriate architecture and weights. 

Given a CPWL predictor, the output for each sample will be uniquely determined by a single line segment. No samples will be predicted by the introduced auxiliary line segments, thus they have no effect on the empirical risk. Let $ f^\ast $ be the CPWL function composed of linear pieces $ f^\ast_1, \cdots, f^\ast_P $, then the corresponding empirical risk is 
\begin{equation} \label{eq4_1}
	R(f^\ast)= \frac{1}{N} \sum_{j=1}^{P}\sum_{i=1}^{N_j} l(f^\ast_j(x_i), y_i).
\end{equation}
Since each optimal line segment $ f^\ast_j $ minimizes the MSE loss of each group of samples (see \eqref{eq4_0}), the total loss $ R(f) $ is also minimized by $ f^\ast $. When perturbating the line segments around the optimal ones, the CPWL predictor $f $ changes correspondingly. However, if the perturbation is small enough, each sample is still predicted by the same linear piece (with perturbation) as before. Use $ f^\prime_j $ to denote the $ j $th line segment after perturbation and $ f^\prime $ to denote the resulted CPWL predictor,   after small perturbation the empirical loss becomes 
\begin{equation} \label{eq4_2}
	R (f^\prime) = \frac{1}{N} \sum_{j=1}^{P}\sum_{i=1}^{N_j} l(f^\prime_j(x_i), y_i),
\end{equation}
where the expression of empirical risk $ R $ is the same as that in \eqref{eq4_1}. By the fact $ R(f) $ is locally minimized by $ f^\ast $, in a small neighborhood around $ f^\ast $ we have
\begin{equation} \label{eq4_3}
	R (f^\prime) \geq R(f^\ast).
\end{equation}
In other words, the CPWL predictor $ f^\ast $, obtained by fitting line segments to different disjoint group of samples, is a local minimum of empirical loss. CPWL predictor $ f^\ast $ can be produced by ReLU networks with appropriate architecture and weights (described in detail in section \ref{section6}), which means that the weights producing $ f^\ast $ are local minima for such networks. 

It is easy to see these local minima are usually spurious. One can partition the samples into more disjoint groups, and fit each group linearly by minimizing the MSE error as before. When the dataset cannot be fitted perfectly by a linear predictor, fitting with more line segments can often reduce the fitting error. Consequently, $ \left \{ f^\ast_j \ | \ j\in [P^\prime] \right \} $ with $ P^\prime > P $ (not counting the auxiliary segments) may be local minima with lower empirical losses than $ \left \{ f^\ast_j \ | \  j\in [P] \right \} $, hence the weights of ReLU networks producing $ \left \{ f^\ast_j \ | \  j\in [P] \right \} $ may be spurious local minima. Fig. 1(b) and fig. 1(c) show respectively the cases of partitioning samples into three and four disjoint groups. There are many possible partition and fitting patterns, therefore spurious local minima are common as long as finer-grained fittings produce lower empirical losses.  

From this simple example, several key observations can be made. 1. The reason why spurious local minima exist in ReLU networks is as follows: ReLU networks output CPWL functions to fit discrete samples and different fitting patterns usually yield different fitting errors.  These facts are also true for networks with any continuous piecewise linear activation functions (such as leaky ReLU) which still generate piecewise linear predictors. 
2. The idea of regarding network training as fitting samples with CPWL predictors applies to networks and datasets of any size. On one hand, the empirical loss can be zero for over-parameterized networks where sufficient number of linear pieces are produced and each sample can be fitted by a single linear piece with zero loss. The redundant linear pieces will not correspond to any data sample and act like auxiliary pieces. However, spurious local minima can still exist for over-parameterized networks as long as it is possible to fit a group of samples by a single linear piece with nonzero loss. On the other hand, if the number of samples is small enough such that a single linear piece can fit them perfectly, the empirical loss is zero and thus no spurious local minima exist. For example, a dataset of two samples in any input dimension will not produce spurious local minima.   

Although in the above example a regression problem is analyzed with MSE loss and one-dimensional input and output, the principle behind is general and applicable to networks with arbitrary continuous loss functions and piecewise linear activations, and higher-dimensional input and output. The general results and their proofs will be given in subsequent sections.

\begin{figure*}
	\centering
	\subfigure[fit with two line segments]{
		\label{} 
		\includegraphics[width=3.5cm]{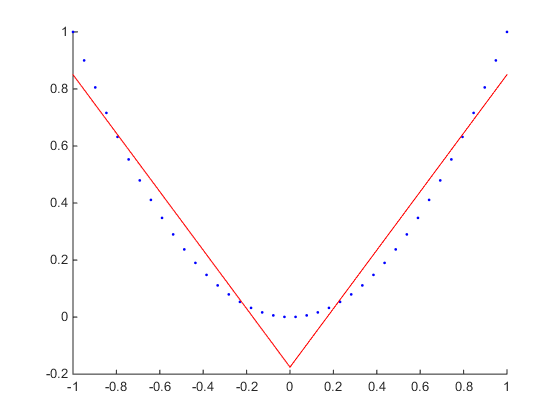}}
	\hspace{1cm}
	\subfigure[fit with three line segments and use an auxiliary segment (the shortest one) to form a continuous piecewise linear predictor]{
		\label{} 
		\includegraphics[width=3.5cm]{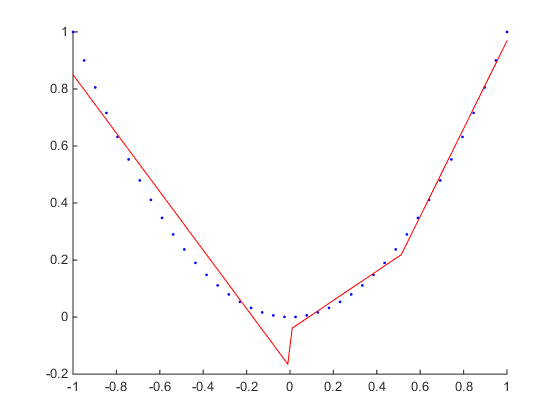}}
	\hspace{1cm}
	\subfigure[fit with four line segments]{
		\label{} 
		\includegraphics[width=3.5cm]{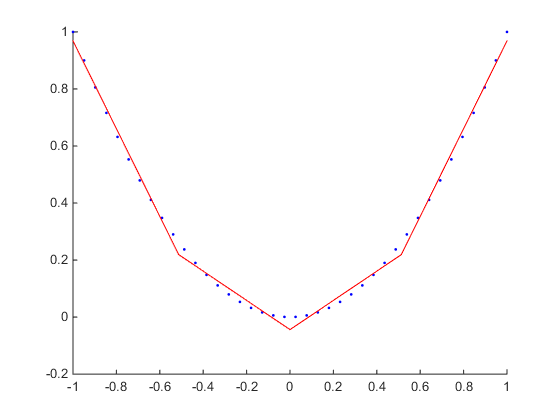}}
	\caption{A motivating example: fitting the same set of data samples with different patterns leads to spurious local minima.}
	\label{fig:twopicture} 
\end{figure*}

\section{Main Results}\label{section5}
In this section, the assumptions and main theorems are given, along with the proof sketch of main theorems.

\subsection{Assumptions}\label{section5.1}

The main theorems of this paper are based on the following assumption.

\textbf{Assumption 1 (data):} \textit{The training data cannot be fitted by a linear model.  i.e., for $ \mathbf{\hat{y}}_i = W \mathbf{x}_i +  \mathbf{b} $, there is $ \sum_{i=1}^{N} l(\mathbf{\hat{y}}_i, \mathbf{y}_i) \neq 0 , \ \forall (W,\mathbf{b}) $}.


%


The reason why this assumption is necessary is that for dataset that can be fitted by linear models, any subset of data samples can also be fitted by linear models and has a zero empirical loss due to the non-negativeness of loss function, thus no spurious local minima exist.  Practical data such as image set usually cannot be fitted by linear models, therefore this assumption is reasonable.



\subsection{Theorems}\label{section5.2}

The main results of this paper are summarized in the following two theorems. Theorem 1 is for deep fully-connected networks and theorem 2 is for CNNs.

\newtheorem{lemma}{Lemma}
\newtheorem{thm}{Theorem}

\begin{thm}\label{theorem1}	
	For dataset satisfying Assumption 1 and deep fully-connected ReLU networks, the empirical risk has spurious local minima for networks with as few as 2 hidden layers and $ 2 d_y $ neurons in each hidden layer. Moreover, the number of non-homogeneous spurious local minima grows for bigger deep fully-connected ReLU networks.  
\end{thm}

\begin{thm}\label{theorem2}	
	For dataset satisfying Assumption 1 and CNNs, the empirical risk has spurious local minima for networks with as few as 2 fully-connected hidden layers and $ 2 d_y $ neurons in each fully-connected hidden layer, and $ 2 d_y $ filters in each convolutional layer. Moreover, the number of non-homogeneous spurious local minima grows for bigger CNNs.  
	
\end{thm}

In these theorems, homogeneousness means that multiplying a constant $ a \neq 0 $ with the weights of one layer and multiplying  $ \frac{1}{a} $ with the weights of an adjacent layer will keep the empirical risk constant, and  thus giving a spurious local minimum, one can scale its weights homogeneously to get another spurious local minimum. Such homogeneousness is excluded when counting the number of spurious local minima in theorem 1 and theorem 2. 


The detailed constructions of network weights that are spurious local minima will be given in the following proof sketch and the complete proofs will appear in section \ref{section6}. It should be pointed out that although the explicit constructions in this paper require a width of $ 2 d_y $ by producing each linear piece independently, spurious local minima  exist for much smaller deep ReLU networks when $ d_y \gg 1 $ since composition of layers can also produce locally optimal CPWL predictors (see Remark 1  in subsection \ref{section6.2} for more explanations).

\subsection{Proof Sketch}\label{section5.3}

In principle, networks with any CPWL activation functions will output CPWL predictors, and when using such CPWL predictors to fit data samples, the existence of different fitting patterns leads to spurious local minima. Therefore, any network architecture and associated weights that can produce locally optimal CPWL predictors are evidences of existing spurious local minima. However, in order to be intuitive,  following the idea given in section \ref{section4}, theorem 1 and theorem 2 will be proved by constructing easy to understand deep ReLU networks and CNNs that give rise to spurious local minima.

This subsection will give the proof sketch and the detailed proofs will be presented in section \ref{section6}.

The key proof steps and their rough ideas are as follows.

\textbf{1. Partition the domain into disjoint convex polytopes.} Partition the domain of interest that contains all samples $ \left \{ \mathbf{x}_i \in   \mathbb{R}^{d_x}, \ i \in [N] \right \}  $ into some disjoint convex polytopes. Data samples are then divided into disjoint groups by these polytopes. Such partition is always possible, and since there are gaps between neighboring samples, one can always construct partitions where there exist intermediate regions between adjacent groups of samples in which  no samples reside. 

\textbf{2. Fit each group of samples with a linear predictor.} The intersections of adjacent linear predictors are convex. If the intersections do not lie in the intermediate regions between adjacent groups of samples, the intermediate regions are then further partitioned into finer polytopes and a auxiliary linear piece is introduced for each finer polytope in the intermediate regions, like the auxiliary line segment in fig.1 (b). The purpose of introducing such auxiliary linear pieces is to form a continuous piecewise linear predictor. All samples in a group will be predicted by the same linear piece. 

\textbf{3. Formulate the CPWL predictor as a maximum over minimum of affine functions.}  It has been shown \cite{CPWL} that any  CPWL function can be formulated as $ \text{max}_i (\text{min}_{j=1}^{K_i} f_j (\mathbf{x})) $, where $ f_j $ is the $ j $th linear piece of CPWL predictor, $ K_i $ is the cardinality of a set of linear pieces.  

\textbf{4. Construct a deep  fully-connected ReLU network or a CNN to produce the CPWL predictor.} By designing appropriate network architectures and weights, the desired CPWL predictors can be generated in the form of $ \text{max}_i (\text{min}_{j=1}^{K_i} f_j (\mathbf{x})) $. 

\textbf{5. Show that weights producing desired CPWL predictors are local minima.} This can be seen by showing that the small perturbation of network weights does not change the fitting pattern, i.e., each sample is still predicted by the original linear piece (with perturbation) responsible for it, and by the fact that the empirical loss of each group of samples has been minimized by corresponding linear piece in step 2. 

\textbf{6. Show that local minima producing CPWL predictors are spurious.} Take a group of samples with nonzero empirical risk and choose a sample in it with nonzero loss, and further divide the region of this group into some finer convex polytopes such that only the single chosen sample is located in a specific polytope. By fitting this sample with a single linear piece  and thereby reducing its loss to zero, and fitting optimally other groups of samples in finer polytopes, the total empirical risk is decreased. In other words, there exist finer partitions of samples with lower empirical risks, thus the local minima resulted from coarser partitions are spurious.

\section{Proofs}\label{section6}

In this section, following the proof sketch in subsection \ref{section5.3}, the detailed proofs are presented. The proofs are general and apply to any continuous loss function, any dataset satisfying assumption 1 and network architecture meeting the size requirements in theorem 1 and theorem 2.


\subsection{Deep fully-connected ReLU networks}\label{section6.1}

\subsubsection{Partition data samples into disjoint groups}

Let us partition the domain of interest $ D $ into some disjoint convex polytopes $ \left \{ R_i, i \in [P] \right \}   $ such that $ D=\bigcup^P_{i=1} R_i $ and $ R_i \bigcap R_j = \varnothing , \  i \neq j $, where $ P $ is the number of partitions. The samples are correspondingly divided into groups and samples in each group are located in the same polytope. It is required that there exist intermediate empty (no samples)  regions between adjacent groups of samples. Such partition always exists since there are gaps between data points. It is possible that some polytopes contain no samples.  

\subsubsection{Fit each each group of samples with a linear predictor}

The samples in the same polytope are then fitted using a linear predictor. Formally, the optimal linear predictor $ \mathbf{f}^\ast_i $ for the $ i $th group of samples is given by 
\begin{equation} \label{eq6_2}
	\mathbf{f}^\ast_i=\text{arg min}_{(A, b)} \sum_{\mathbf{x}_k \in R_i } l( \mathbf{f}_i (\mathbf{x}_k), \mathbf{y}_{k}), \ \  i \in [P],
\end{equation}
where $ \mathbf{f}_i (\mathbf{x}) = A \mathbf{x} + \mathbf{b} \in \mathbb{R}^{d_y} $ is a linear function of input $ \mathbf{x} $ with $ A \in \mathbb{R}^{d_y \times d_x} $ and $ \mathbf{b} \in \mathbb{R}^{d_y} $, region $ R_i $ is the polytope corresponding to the $ i $th group of samples.  
$ \mathbf{f}^\ast_i $ always exists for any loss function $ l $.


If the intersections (which are convex) of adjacent optimal linear predictors are located outside the corresponding intermediate regions, in order to form a proper CPWL predictor, the involved adjacent polytopes are shrunk slightly such that the grouping of samples does not change.  Then, the empty intermediate regions between shrunk polytopes are further partitioned into some auxiliary polytopes (such as partition into stripes of simplex). In each auxiliary polytope, a linear piece is constructed that connects the linear pieces in adjacent polytopes such that a CPWL function is formed. There are no samples in auxiliary polytopes,  the total empirical loss is thus not affected by them.

Define $ \mathbf{f}^\ast $ as the  CPWL function composed of linear pieces $ \mathbf{f}^\ast_i \ (i \in [P]) $ of different groups of samples (including auxiliary pieces), i.e.,  $ \mathbf{f}^\ast = \mathbf{f}^\ast_i $ in the $ i $th region, the total empirical risk in then given by 
\begin{equation} \label{eq6_3}
	R(\mathbf{f}^\ast)=\frac{1}{N} \sum_{i=1}^{P}\sum_{k=1}^{N_i} l( \mathbf{f}^\ast_i (\mathbf{x}_k), \mathbf{y}_k).
\end{equation}

\subsubsection{Formulate the CPWL predictor as a maximum over minimum of linear pieces}

Given a scalar CPWL function $ f $ and a set of linear functions $ \left \{  f_i , \  i\in [P]  \right \} $, suppose there are  $ P $ regions $ \left \{  R_i \right \} $ in the domain such that in the $ i $th region, there is $ f=f_i \  (i\in [P]) $, i.e., each region corresponds to a single linear piece.  
By \cite{CPWL}, any CPWL function $ f $  can be formulated by $ f =  \text{max}_i (\text{min}_{j=1}^{K_i} f_j) $, where $ f_j $ is the $ j $th linear piece. More specifically, 
\begin{equation} \label{eq6_4}
	f = \text{max}^{P}_{i=1} \Psi _i ,
\end{equation}
with
\begin{equation} \label{eq6_5}
	\Psi _i = \text{min}_{j=1}^{K_i} f_j , \  \forall f_j \ | \ f_j \geq f_i \ \text{in every point of region} \ R_i.  
\end{equation}
$ K_i $ is the number of affine functions $ f_j $ that satisfy $ f_j \geq f_i \ \text{in every point of} \ R_i $. 

The process of computing $ f $ can be illustrated by the example shown in fig. 2. There are three affine functions in fig. 2: $ f_1 $, $ f_2 $ and $ f_3 $, and the blue thick line segments represent the CPWL function $ f $. The real domain is divided into three disjoint regions: $ R_1 $, $ R_2 $ and $ R_3 $, and in each region the CPWL function equals to a specific affine function. $ \Psi _1 $ corresponding to region $ R_1 $ is $ \Psi _1 = \text{min} (f_1,f_2)  $;  similarly, $ \Psi _2 = \text{min} (f_1,f_2)  $ and $ \Psi _3 = f_3  $. Consequently, $ \Psi _1 = f_1$ in region $ R_1 $, $ \Psi _1 = f_2$ in region $ R_2 $ and $ \Psi _1 = f_2$ in region $ R_3 $. Note that $ \Psi _i $ may vary from region to region. Similarly, $ \Psi _2 = f_1$ in region $ R_1 $, $ \Psi _2 = f_2$ in region $ R_2 $ and $ \Psi _2 = f_2$ in region $ R_3 $.  $ \Psi _3 = f_3$ in all regions. Then, by \eqref{eq6_4}, $ f = \text{max}^{3}_{i=1} \Psi _i $, so $ f = \text{max} (f_1,f_3)=f_1$ in region $ R_1 $, $ f =\text{max} (f_2,f_3) = f_2$ in region $ R_2 $,  and  $ f = \text{max} (f_2,f_3) = f_3$ in region $ R_3 $. This is exactly the desired CPWL function shown in blue thick line segments. Interested readers are referred to \cite{CPWL} for the detailed proof.

\begin{figure}[ht]
	\vskip 0.2in
	\begin{center}
		\centerline{\includegraphics[width=5.0cm]{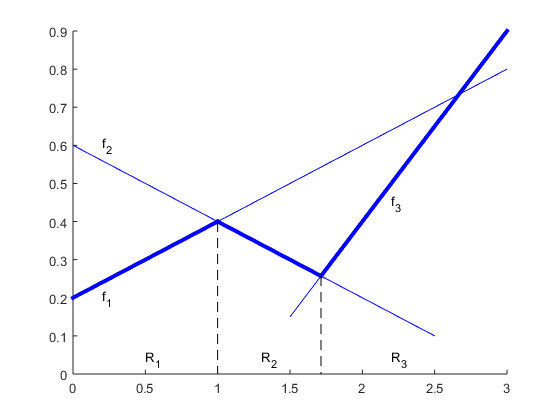}}
		\caption{A continuous piecewise linear function can be expressed as a maximum over minimum of a set of linear functions.}
		\label{celldef}
	\end{center}
	\vskip -0.2in
\end{figure}

\subsubsection{Construct a deep ReLU network to produce the CPWL predictor}

Given any dataset satisfying assumption 1, a deep ReLU network with appropriate architecture and parameters is designed in this step to produce the CPWL function formulated in \eqref{eq6_4} and \eqref{eq6_5}. Since the method of representing a CPWL predictor as a maximum over minimum of linear pieces in step 3 only applies to scalar functions, so each component of optimal CPWL predictor $ \mathbf{f}^\ast $ will be produced independently at first and then concatenated. In the following,  deep ReLU networks producing scalar CPWL predictors will be mainly described.

Given an optimal scalar CPWL function $ f^\ast $ and its linear pieces $ \left \{  f^\ast_i, \ i \in [P] \right \} $, each linear piece $ f^\ast_i $  can be produced by a single ReLU neuron that outputs $ \sigma (\mathbf{w}^T_i \mathbf{x} + b_i) $, where weight vector $ \mathbf{w}_i $ and bias $ b_i $ are parameters of $ f^\ast_i $. These ReLU neurons producing linear pieces $ \left \{  f^\ast_i \ (i \in [P]) \right \} $ are put at the first hidden layer that directly connects to input $ \mathbf{x} $. Since ReLU activation function truncates negative input, in order to preserve the shape of linear pieces, a large positive constant bias $ c $  is added to the input of each  ReLU neuron in the first hidden layer such that its output is positive in the domain of interest. Then, each ReLU neuron in the first hidden layer outputs a linear piece in the domain of interest $ D $,
\begin{equation} \label{eq6_6}
	f^\ast_i = \sigma (\mathbf{w}^T_i \mathbf{x} + b_i + c) = \mathbf{w}^T_i \mathbf{x} + b_i + c > 0, \ \mathbf{x} \in D, \ i \in [P].
\end{equation}
The effect of this large positive constant bias will be cancelled at the output layer by subtracting $ c $ from each output component.

One needs to implement the maximum and minimum operations using ReLU networks. 
Minimization can be written as 
\begin{equation} \label{eq6_7}
	\text{min} (a,b) = b - \text{max}(0,b-a)= \sigma(\sigma(b) - \sigma(b-a)), \ \ a, b \geq 0 ,
\end{equation}
thus each minimization can be represented by a 2-layer ReLU network  with 2 hidden neurons and specific weights shown in fig. 3(a). The minimization over $ K_i>2 $ affine functions can be implemented hierarchically, where the minimizations for groups of two affine functions are obtained at first, and then intermediate results are grouped likewise to compute  minimizations at the next level, until finally $ \text{min}_{j=1}^{K_i} f_j $ is obtained. Fig. 3(c) shows the hierarchical architecture used to compute the minimization over four affine functions. 

The maximum over two variables can be written as 
\begin{equation} \label{eq6_8}
	\text{max} (a,b) = a + \text{max}(0,b-a)= \sigma (\sigma(a) + \sigma(b-a) ), \ \ a, b \geq 0. 
\end{equation}
Similarly, $ \text{max}^{P}_{i=1} \Psi _i $ can be implemented hierarchically, until finally each neuron in the output layer produces a desired CPWL predictor.  For fully-connected networks, the connections not involved in these computations are associated with zero weight vectors and biases.

\begin{figure*}
	\centering
	\subfigure[minimization]{
		\label{} 
		\includegraphics[width=3.0cm]{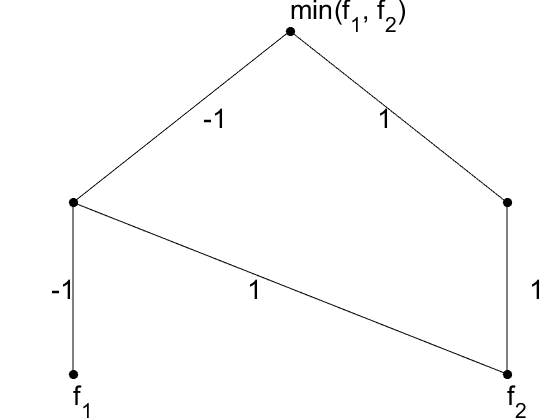}}
	\hspace{1cm}
	\subfigure[maximization]{
		\label{} 
		\includegraphics[width=3.0cm]{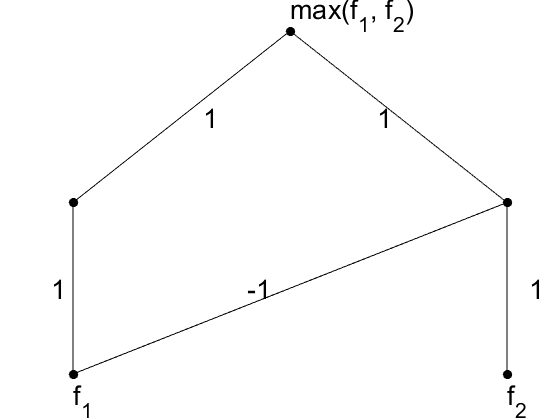}}
	\hspace{1cm}
	\subfigure[hierarchical implementation of minimization over four affine functions]{
		\label{} 
		\includegraphics[width=4.0cm]{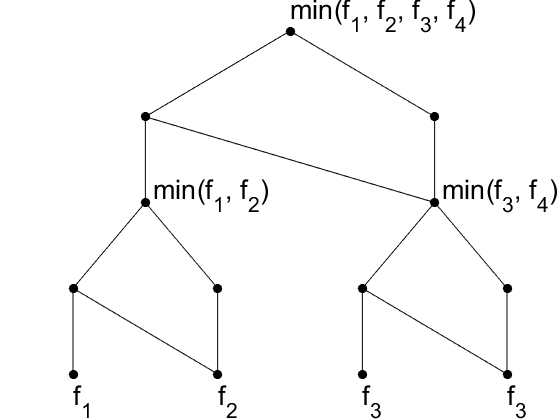}}
	\caption{ReLU network architectures used to represent maximization and minimization operations.}
	\label{fig:twopicture} 
\end{figure*}

\subsubsection{Show that weights producing CPWL predictors are local minima}

$ \left \{  \mathbf{f}^\ast_i, \ i \in [P] \right \}  $ minimizes the empirical risk of each group of samples. In the following, it will be shown that there exist a small neighborhood around the weights producing $ \mathbf{f}^\ast $ in which the fitting pattern does not change, i.e., each data sample is still fitted by the linear piece originally responsible for it  when perturbating networks weights in this neighborhood. 

ReLU network architectures have been presented in step 4 to produce each component of $ \mathbf{f}^\ast $ independently. By arranging the components of  $ \mathbf{f}^\ast $ sequentially, the output of network producing $ \mathbf{f}^\ast $ can be expressed as
\begin{equation} \label{eq6_10}
	\hat{\mathbf{y}}^{i}=W^{L} I^{L-1, i} W^{L-1} I^{L-2, i} W^{L-2} \cdots W^{1} \mathbf{x}_{i} .
\end{equation}
Here the biases have been absorbed in weights. The weight matrix $ W^1$ in the first layer specifies the linear pieces $ \left \{  f^\ast_{i,j}, \ i \in [P], j \in [d_y] \right \}  $ for different regions and components, where $ f^\ast_{i,j} $ is the $ j $th component of $ \mathbf{f}^\ast _i $. The layers from 2 to $ L $ with associated weight matrix $ W^i \ (i=2,3,...,L ) $ implement the minimization and maximization operations of each component  used to produce the CPWL predictor. 
$ \left \{ I^{l, i}, \ l=2,3,\cdots,L-1 \right \}  $ are determined by the results of these minimization and maximization operations, and some ReLU neurons in layers from 2 to $ L-1 $ may be inactive. 

Given fixed weights,
output $ \hat{\mathbf{y}}^{i} $ will be a linear function of input $ \mathbf{x}_{i} $ if $ \left \{ I^{l, i}, \ l\in [L-1] \right \}  $ remain constant. The samples in each group are predicted by the same linear predictor and hence they share the same activation pattern, i.e., $ \left \{  I^{l, i}, \ l\in [L-1] \right \}  $ are identical for all samples $ \left \{   \mathbf{x}_{i} \right \} $ in the same group and so their outputs $ \left \{ \hat{\mathbf{y}}^{i}  \right \}  $  have the same linear expressions. 
Given optimal linear predictors $ \left \{  \mathbf{f}^\ast_i, \ i \in [P] \right \}  $ and the fact that $ \mathbf{f}^\ast (\mathbf{x}_i) = \mathbf{f}^\ast_j( \mathbf{x}_i) = W^{L} I^{L-1, i} W^{L-1} I^{L-2, i} W^{L-2} \cdots W^{1} \mathbf{x}_i $ for each sample $ \mathbf{x}_i \in R_j $, the empirical risk can be written as
\begin{equation}\label{eq6_12}
	R(W^{1}, \cdots, W^{L} ) = \frac{1}{N} \sum_{j=1}^{P}\sum_{i=1}^{N_j} l(\mathbf{f}^\ast_j(\mathbf{x}_i), \mathbf{y}_i) .
\end{equation}

When perturbating the weights producing $ \mathbf{f}^\ast $ as $ W^{\prime l}=W^{l}+\delta W^{l} \ (l\in [L])$, the linear predictors $ \left \{ \mathbf{f}_j \right \}  $ and activation pattern $ \left \{ I^{l, i} \ (i\in [N], \ l\in [L-1])  \right \}  $ will be perturbated correspondingly. However, if the perturbation is small enough, $ \left \{ I^{l, i} \ (i\in [N], \ l\in [L-1])  \right \}  $ will keep constant. Let $ \epsilon = \text{min}_{(l,m,n)}  | \delta W^{l} (m,n) | $ such that $ \left \{ I^{l, i} \ (i\in [N], \ l\in [L-1])  \right \}  $ keep constant after perturbation, then the form of linear predictors  $ \left \{ \mathbf{f}_j , \ j\in [P] \right \}  $ does not change (except the perturbation of weights themselves) after perturbations $ W^{\prime l}=W^{l}+\delta W^{l} \ (l\in [L])$ satisfying  $ \text{max}_{(l,m,n)}  | \delta W^{l} (m,n) | \leq \epsilon$,
\begin{equation}\label{eq6_13}
	\begin{aligned}
		\mathbf{f}^\prime_j (\mathbf{x}_{i}) &=(W^{L}+\delta W^{L}) I^{L-1, i} (W^{L-1}+\delta W^{L-1})  \\
		& I^{L-2, i}  (W^{L-2}+\delta W^{L-2}) \cdots (W^{1}+\delta W^{1}) \mathbf{x}_{i}, \qquad \mathbf{x}_{i} \in R_j, \ j\in [P] ,
	\end{aligned}
\end{equation}
After rearrangement, 
\begin{equation}\label{eq6_14}
	\begin{aligned}
		\mathbf{f}^\prime_j (\mathbf{x}_{i}) &= W^{L} I^{L-1, i} W^{L-1} I^{L-2, i} W^{L-2} \cdots W^{1} \mathbf{x}_{i}  \\
		& + \delta W^{L} I^{L-1, i} W^{L-1} I^{L-2, i} W^{L-2}\cdots W^{1} \mathbf{x}_{i}  \\
		& +  W^{L} I^{L-1, i} \delta W^{L-1} I^{L-2, i} W^{L-2}\cdots W^{1} \mathbf{x}_{i} + \cdots \\
		& + \delta W^{L} I^{L-1, i} \delta W^{L-1}  I^{L-2, i} \delta W^{L-2}\cdots \delta W^{1}  \mathbf{x}_{i} \\
		&=  \mathbf{f}^\ast_j (\mathbf{x}_{i}) + \delta \mathbf{f}_j (\mathbf{x}_{i}) ,\ \mathbf{x}_{i} \in R_j, \ j\in [P],
	\end{aligned}
\end{equation}
where $ \delta \mathbf{f}_j (\mathbf{x}_{i}) = [\delta W^{L} I^{L-1, i} W^{L-1} I^{L-2, i} W^{L-2}\cdots W^{1}  +  W^{L} I^{L-1, i} \delta W^{L-1} I^{L-2, i} W^{L-2}\cdots W^{1} + \cdots + \delta W^{L} I^{L-1, i} \delta W^{L-1}  I^{L-2, i} \delta W^{L-2}\cdots \delta W^{1} ] \mathbf{x}_{i} $. The form of
$ \delta \mathbf{f}_j $ is the same for all samples in a group since they share the same activation pattern $ \left \{  I^{l, i} \ (\mathbf{x}_{i} \in R_j, \ l\in [L-1]) \right \}  $, and consequently these samples still share the same linear predictor after perturbation. $ \mathbf{f}^\ast_j (\mathbf{x}) + \delta \mathbf{f}_j (\mathbf{x}) $ is still a linear function of $ \mathbf{x} \in R_j  $.

The empirical risk after perturbation is given by
\begin{equation}\label{eq6_15}
	R(W^{^\prime 1}, \cdots, W^{^\prime L} ) = \frac{1}{N} \sum_{j=1}^{P}\sum_{i=1}^{N_j} l(\mathbf{f}_j^\prime (\mathbf{x}_i), \mathbf{y}_i) .
\end{equation}
However, since $ \mathbf{f}^\ast_j $ minimizes $ \sum_{i=1}^{N_j} l( \mathbf{f}_j (\mathbf{x}_i), \mathbf{y}_i) $, there exist a small neighborhood around $ \left \{ W^{l}, \ l\in [L]  \right \} $ such that    for all points $ \left \{ W^{^\prime l}, \ l\in [L]  \right \} $ satisfying  $ \text{max}_{(l,m,n)} | W^{^\prime l} (m,n) -  W^{l}  (m,n) | < \epsilon \prime $, there is
\begin{equation}\label{eq6_16}
	\sum_{i=1}^{N_j} l( \mathbf{f}^\ast_j (\mathbf{x}_i), \mathbf{y}_i)  \leq \sum_{i=1}^{N_j} l( \mathbf{f}^\ast_j (\mathbf{x}_{i}) + \delta \mathbf{f}_j (\mathbf{x}_{i}), \mathbf{y}_i)  ,\  j\in [G] .
\end{equation}
Consequently,  we have
\begin{equation}\label{eq6_17}
	R(W^{1}, \cdots, W^{L} ) \leq R(W^{^\prime 1}, \cdots, W^{^\prime L} )
\end{equation}
in a small neighborhood around $ \left \{ W^{l}, \ l\in [L]  \right \} $ with  $ \text{max}_{\left \{ l,m,n \right \} } | \delta W^{l} (m,n) | < \text{min} (\epsilon, \epsilon \prime) $. Therefore, $ W=\left \{ W^{l}, \ l\in [L]  \right \} $ is a local minimum for the networks constructed in step 4.

\subsubsection{Show that local minima producing CPWL predictors are often spurious}

In this step, it will be shown that the local minima $ \left \{ W^{l}, \ l\in [L]  \right \} $ constructed in step 4 are usually spurious. Take a group of samples with nonzero empirical risk (without loss of generality, suppose it is the $ k $th group), i.e., $ \sum_{i=1}^{N_k} l(\mathbf{f}_k(\mathbf{x}_i), \mathbf{y}_i) \neq 0 $. By assumption 1, such group at least always exist when fitting all samples with a single linear predictor. Due to the non-negative nature of loss function, there must be some samples in the $ k $th group with nonzero loss. Pick any sample $ \mathbf{x}_n $ with nonzero loss in the $ k $th group,   
the empirical risk can then be expanded as
\begin{equation}\label{eq6_18}
	R = \frac{1}{N} [ \sum_{j=1, j\neq k}^{P}\sum_{i=1}^{N_j} l( \mathbf{f}_j(\mathbf{x}_i), \mathbf{y}_i ) + \sum_{i=1, \ i\neq n}^{N_k} l(\mathbf{f}_k (\mathbf{x}_i), \mathbf{y}_i ) +  l(\mathbf{f}_k(\mathbf{x}_n), \mathbf{y}_n ) ].
\end{equation}

Let us further partition the region $ R_k $  into some  disjoint convex subregions, such that a subregion contains the single point $ \mathbf{x}_n $ and other samples are contained in remaining subregions. Regions other than $ R_k $ keep intact.
The minimal possible value of loss function $ l $ is zero, so there exists $ \hat{\mathbf{y}}_n $ such that $ l(\hat{\mathbf{y}}_n, \mathbf{y}_n)=0 $. Then, the single sample $ \mathbf{x}_n $ can be fitted by a linear predictor $ \hat{\mathbf{f}}(\mathbf{x}_n)= \hat{A}\mathbf{x}_n + \hat{\mathbf{b}} $,  where $ \hat{A} \in \mathbb{R}^{d_y \times d_x} $ and $ \hat{\mathbf{b}} \in \mathbb{R}^{d_y} $ are parameters of $ \hat{\mathbf{f}} $. Because there always exists $ (\hat{A},\hat{\mathbf{b}}) $ such that $ \hat{A}\mathbf{x}_n + \hat{\mathbf{b}} = \hat{\mathbf{y}}_n $, there is
\begin{equation}\label{eq6_19}
	l(\hat{\mathbf{f}}(\mathbf{x}_n), \mathbf{y}_n) = 0 .
\end{equation}
The remaining samples in region $ R_k $ have been partitioned into subgroups and each subgroup is again fitted optimally by linear predictors. Suppose for the $ j $th subgroup (excluding the one containing $ \mathbf{x}_n $) of region $ R_k $,  the new optimal linear predictor is $ \hat{\mathbf{f}}_{kj} $ and there are $ N_{kj} $ samples in it, then 
\begin{equation}\label{eq6_20}
	\sum_{i=1}^{N_{kj}} l(\hat{\mathbf{f}}_{kj} (\mathbf{x}_i), \mathbf{y}_i) \leq \sum_{i=1}^{N_{kj}} l(\mathbf{f}_{k} (\mathbf{x}_i), \mathbf{y}_i) 
\end{equation} 
due to the fact that $ \hat{\mathbf{f}}_{kj} $ is optimal for the $ j $th subgroup of samples in region $ R_k $.

At last, combination of \eqref{eq6_18},\eqref{eq6_19} and \eqref{eq6_20} yields
\begin{equation}\label{eq6_21}
	\begin{aligned}
		\hat{R} &= \frac{1}{N} \sum_{j=1, j\neq k}^{P}\sum_{i=1}^{N_j} l(\mathbf{f}_j(\mathbf{x}_i), \mathbf{y}_i) + \frac{1}{N} \sum_{j} \sum_{i=1}^{N_{kj}} l(\hat{\mathbf{f}}_{kj} (\mathbf{x}_i), \mathbf{y}_i) + \frac{1}{N} l(\hat{\mathbf{f}} (\mathbf{x}_n), \mathbf{y}_n) 
		\\
		& < \frac{1}{N} \sum_{j=1, j\neq k}^{P}\sum_{i=1}^{N_j} l(\mathbf{f}_j(\mathbf{x}_i), \mathbf{y}_i) +  \frac{1}{N} \sum_{j} \sum_{i=1}^{N_{kj}} l(\mathbf{f}_{k} (\mathbf{x}_i), \mathbf{y}_i)  +  \frac{1}{N} l(\mathbf{f}_k(\mathbf{x}_n), \mathbf{y}_n) = R, 
	\end{aligned}
\end{equation} 
where the fact $  l(\mathbf{f}_k(\mathbf{x}_n), \mathbf{y}_n)  > 0 $ has been used for the inequality. 
This indicates that lower empirical risks may be obtained with finer partitions, hence $ \left \{ W^{l}, \ l\in [L]  \right \} $ is a spurious local minimum. Whether finer partitions with lower empirical risks exist depends on data and current partition of samples.

\subsubsection{Size of deep fully-connected ReLU networks with spurious local minima}

For a CPWL fitting with only two linear pieces $ \mathbf{f}_1 $ and $ \mathbf{f}_2 $, at most one minimization or maximization operation is required for each component of $ \mathbf{f}_1 $ and $ \mathbf{f}_2 $, thus a network with 2 hidden layers (one is for producing affine functions $ \mathbf{f}_1 $ and $ \mathbf{f}_2 $, the other is for minimization or maximization of each component of $ \mathbf{f}_1 $ and $ \mathbf{f}_2 $) and $ 2 d_y $ neurons in each hidden layer (for each component, the width of minimization or maximization module is 2) is enough to represent this CPWL function. For data samples satisfying assumption 1, i.e., they cannot be fitted by linear models, comparing with fitting by a single linear predictor (which apparently can be produced by the same network), there always exist two-piece CPWL fittings with lower empirical risks. Therefore, spurious local minima exist for deep ReLU networks with as few as 2 hidden layers and a width of $ 2 d_y $ neurons and datasets that cannot be linearly fitted. 

For a CPWL fitting with $ n \ (n>2)$ linear pieces, there are at most $ n $ affine functions in each $ \Psi _i $, thus the network can include at most $ 4 \lfloor \text{log}_2 n \rfloor $ hidden layers, with $ 2 \lfloor \text{log}_2 n \rfloor $ layers for hierarchical minimization required by each $ \Psi _i $ and another $ 2 \lfloor \text{log}_2 n \rfloor $ layers for hierarchical maximization over all $ \Psi _i \ (i \in [n])$. The width of hidden layers is $ n^2 d_y $, due to that minimization requires at most $ n $ neurons in a hidden layer for each $ \Psi _i  \ (i \in [n])$  and each component. 


For bigger deep fully-connected ReLU networks, more finer partitions and fittings can be generated, and consequently more spurious local minima exist, leading to their prevalence. The actual width of networks possessing spurious local minima can be much less than $ n^2 d_y $, see Remark 1 later in this section.

\subsection{CNNs}\label{section6.2}

The main differences of CNNs with fully-connected networks come from the convolutional layers and pooling layers. In this subsection, it will be proved that for CNNs with convolutional layers, pooling layers and fully-connected layers, spurious local minima are common.

The main idea is as follows. The convolution is a linear operation. Let each neuron in the first convolutional layer computes a positive value $  \sigma ( \left \langle  \mathbf{w}^1_t, \mathbf{x}_p  \right \rangle + b^1_h  + c) =  \left \langle  \mathbf{w}^1_t, \mathbf{x}_p  \right \rangle + b^1_h  + c $, where $ \mathbf{w}^1_t $ is the $ t $th convolutional filter in the first convolutional layer and $ \mathbf{x}_p $ is the $ p $th input patch. $ c $ is a large positive constant to ensure a positive output in the domain of interest and its effect is cancelled at the output layer by subtracting $ c $ from each output neuron. The output of the first convolutional layer is thus a linear function of input $ \mathbf{x} $. Average pooling is also a linear operation.

One can partition the data samples into groups as before. For a CNN with stacked convolutional layers, average pooling layers and fully-connected layers, the output is still linear with respect to input of each group in the domain of interest. Therefore, each group of samples can be optimally fitted using a linear predictor producible by CNN, and CNN architectures  can be constructed to generate these optimal linear predictors, along with point-wise maximization and minimization operations over them. Like the case of deep fully-connected networks, under small perturbations of weights, the linear predictors also go through small perturbations, but the activation pattern and consequently the expression of empirical risk for each group of samples remain constant, hence the constructed CNN weights are local minima. By further partitioning a group of samples with nonzero loss into subgroups, it can be shown that these local minima are spurious since finer partitions have lower fitting errors.   

If pooling operation is the max-pooling, since point-wise maximization over linear functions (and CPWL functions) of input $ \mathbf{x} $ generally produces a CPWL function, the final outputs of CNNs are therefore CPWL functions, and each group of samples will then be fitted by these CPWL predictors. Other steps are the same as those for CNNs with average pooling.

The CNN construction for producing the desired CPWL functions is not unique. One construction is given as follows.
The output of a CNN is
\begin{equation} \label{eq6_22}
	\begin{aligned}
		\hat{\mathbf{y}}^{i}= & W^{L} I^{L-1, i} W^{L-1} I^{L-2, i} W^{L-2}\cdots W^{l_{fc}} \\
		& I^{l_{fc}-1, i} W^{l_{fc}-1} * \cdots P \cdots *W^{1} * \mathbf{x}_{i}, 
	\end{aligned}
\end{equation}
where $ * $ is the convolution operation. Layers from $ l_{fc} $ to  $ L $ are fully-connected layers. For convolutional layers, $ I^{l, i} $ is defined for every neuron in the feature maps. $ P $ is the pooling operation, either max pooling or average pooling.  
Given constant weights and constant activation pattern $ \left \{ I^{l, i}  \right \}  $, $ \hat{\mathbf{y}} $ will be either a linear or a CPWL function of input $ \mathbf{x} $, depending on which pooling operation is used. 


The optimal linear or CPWL predictors for different group of samples will be obtained independently.  
Let the linear or CPWL predictor for the $ j $th group of samples be $  \mathbf{f}_j $, with elements being $  f_{j,k} (\mathbf{x}_i) = {\mathbf{w}^{l_{fc}}_{j,k}}^T  (\mathbf{w}^{l_{fc}-1}_{j,k} * \cdots P \cdots * \mathbf{w}^{1}_{j,k} * \mathbf{x}_{i} ), \ (\mathbf{x}_{i} \in R_j, \ j \in [P], \ k \in [d_y] ) $, where $ f_{j,k} $ is the $ k $th component of $  \mathbf{f}_j $, $ \left \{ \mathbf{w}^{l}_{j,k}, \ l \in [l_{fc}-1] \right \} $ are corresponding convolutional filters, and $ \mathbf{w}^{l_{fc}}_{j,k} $ is the corresponding weight vector for the  first fully-connected layer. All weights in  the $ l $th ($ l > 1 $) convolutional layers and the first fully-connected layer are fixed to 1, i.e.,  $ \mathbf{w}^l_{j,k} = (1,1,\cdots,1)^T \in \mathbb{R}^{s_{l-1}} \ (2 \leq l < l_{fc})$ and $ \mathbf{w}^{l_{fc}}_{j,k} = (1,1,\cdots,1)^T  $. The biases in these layers are fixed to 0. Therefore, only the filters and biases in $ \mathbf{w}^{1}_{j,k} $ will be optimized.  The large positive bias $ c $  is then added to the input of each ReLU neuron in the first convolutional layer to guarantee linear output in the domain of interest. The reason why the first fully-connected layer is used in the predictor is that each neuron in the first fully-connected layer has a receptive field as large as the whole image (take images as example), so all pixels are utilized when optimizing the predictors and consequently the network outputs of all images in a group will be perturbated in the same way (like in \eqref{eq6_14}) which guarantees the existence of local minima.

Each optimal linear or CPWL predictor $  \mathbf{f}^\ast _j $ is then independently obtained by minimizing the empirical loss $ \sum_{i=1}^{N_j} l(  \mathbf{f}_j (\mathbf{x}_i), \mathbf{y}_i) ,  \ (\mathbf{x}_i \in R_j, \ j \in [P]) $
through tuning the filters and biases in $ \mathbf{w}^{1}_{j,k} \ (k \in [d_y]) $.
For each convolutional layer and the first fully-connected layer, the filters and biases from different regions and output components  are concatenated to form weight matrix $ W^{l} \ (1 \leq l \leq  l_{fc}) $, i.e., $ W^{l}((k-1) P +j ,:) = {\mathbf{w}^{l}_{j,k}}^T , \ (j \in [P], \ k \in [d_y])$.
Finally, the fully-connected layers above layer $ l_{fc} $ are used for computing maximization over minimization of linear pieces $  f^\ast_{j,k}, $ for each component.
Since  $ \left \{ I^{l, i}, \ 1 \leq l \leq l_{fc} \right \}   $ are identity matrices due to positive inputs of corresponding neurons, we have $ \mathbf{f}^\ast (\mathbf{x}_i) = W^{L} I^{L-1, i} W^{L-1} I^{L-2, i} W^{L-2}\cdots W^{l_{fc}}  I^{l_{fc}-1, i} W^{l_{fc}-1} * \cdots P \cdots *W^{1} * \mathbf{x}_{i}    =  \hat{\mathbf{y}}^{i}  $. 
The CNNs constructed above thus can produce the locally optimal CPWL functions used to fit data samples under different fitting patterns, thus spurious local minima are prevailing for CNNs.

For a dataset partitioned into 2 groups, the above CNN construction with average-pooling needs at least one convolutional layer with $ 2 d_y $ convolutional filters, 2 fully-connected hidden layers (the first fully-connected layer is used for computing optimal linear predictors and the other one for computing $ \text{min} (f_{1,k}, f_{2,k}) $ or $ \text{max} (f_{1,k}, f_{2,k}) $ of each component) with a width of $ 2 d_y $ neurons. For datasets that cannot be fitted by linear models, comparing with fitting all samples by a single linear predictor with nonzero risk, there always exist CPWL fittings with two pieces that have lower empirical risks. Therefore, spurious local minima exist for such CNNs.  For bigger CNNs, finer partitions and fittings can be generated, and consequently more spurious local minima exist. The number of possible fitting patterns depends on data and the number of linear pieces a network can produce.

\textbf{Remark 1:\label{Remark 1}} The width of $ 2 d_y $ in the proposed architecture is obtained by producing the CPWL predictor of each  component  independently. For some typical  CNN architectures, such as AlexNet \cite{AlexNet} and VGG \cite{VGG} for ImageNet classification, this width requirement is satisfied by fully-connected layers (both AlexNet and VGG have 2 fully-connected hidden layers  with a width of 4096). However, it must be pointed out that the width of $ 2 d_y $ is unnecessary in practice, because that output neurons share the lower layers and the parameters of all layers can be jointly optimized to fit each group of samples simultaneously, as long as the network, even much smaller than $ 2 d_y $, is able to express the desired CPWL predictors.

\section{Conclusion}

In this work, the loss landscape of deep fully-connected neural networks and CNNs with piecewise linear activation functions has been studied from a theoretical perspective. It is shown that spurious local minima are common for these networks. The reason for existing spurious local minima is that each output neuron of deep fully-connected networks and CNNs with piecewise linear activations produces a continuous piecewise linear output. When minimizing the empirical risk, different pieces of a continuous piecewise linear output can fit optimally disjoint groups of data samples.  Distinct fitting patterns usually result in different levels of empirical risk, thus spurious local minima are common. Continuous piecewise linear predictors are represented as maximization over minimization of linear pieces, and deep ReLU networks and CNNs are constructed to produce them. 

In future work, the author plans to explore the gap between empirical risks of different spurious local minima, as well as the dynamics of how gradient-based optimization converges to spurious local minima, in the hope of inspiring optimization algorithms that can escape them effectively.

\section*{Acknowledgements}
This work was supported by the National Natural Science Foundation of China [grant number 61806013].

\bibliographystyle{plain}
\bibliography{Understanding_global_landscape_ReLU}

\begin{thebibliography}{10}

\bibitem{Allen-Zhu}
Zeyuan Allen-Zhu, Yuanzhi Li, and Zhao Song.
\newblock A convergence theory for deep learning via over-parameterization.
\newblock In {\em International Conference on Machine Learning}, 2019.

\bibitem{Low-loss-connection}
Ivan Anokhin and Dmitry Yarotsky.
\newblock Low-loss connection of weight vectors: distribution-based approaches.
\newblock In {\em International Conference on Machine Learning}, 2020.

\bibitem{Baldi}
Pierre Baldi and Kurt Hornik.
\newblock Neural networks and principal component analysis: Learning from
  examples without local minima.
\newblock {\em Neural Networks}, 2(1):53–58, 1989.

\bibitem{Bianchini}
M~Bianchini and F~Scarselli.
\newblock On the complexity of neural network classifiers: A comparison between
  shallow and deep architectures.
\newblock {\em IEEE Transactions on Neural Networks and Learning Systems},
  25(8):1553–1565, 2014.

\bibitem{Deforming-Loss-Surface}
Liangming Chen, Long Jin, Xiujuan Du, Shuai Li, and Mei Liu.
\newblock Deforming the loss surface to affect the behaviour of the optimizer.
\newblock In {\em AAAI Conference on Artificial Intelligence}, 2021.

\bibitem{Choromanska15}
Anna Choromanska, Mikael Henaff, Michael Mathieu, Gerard~Ben Arous, and Yann
  LeCun.
\newblock The loss surfaces of multilayer networks.
\newblock In {\em Artificial Intelligence and Statistics}, 2015.

\bibitem{attackingthesaddle}
Yann~N Dauphin, Razvan Pascanu, Caglar Gulcehre, Kyunghyun Cho, Surya Ganguli,
  and Yoshua Bengio.
\newblock Identifying and attacking the saddle point problem in
  high-dimensional non-convex optimization.
\newblock In {\em Advances in Neural Information Processing Systems}, 2014.

\bibitem{Ding}
Tian Ding, Dawei Li, and Ruoyu Sun.
\newblock Spurious local minima exist for almost all over-parameterized neural
  networks.
\newblock {\em Optimization online}, 2019.

\bibitem{EssentiallyNoBarriers}
Felix Draxler, Kambis Veschgini, Manfred Salmhofer, and Fred~A. Hamprecht.
\newblock Essentially no barriers in neural network energy landscape.
\newblock In {\em International Conference on Machine Learning}, 2018.

\bibitem{QuadraticActivation}
Simon~S. Du and Jason~D. Lee.
\newblock On the power of over-parametrization in neural networks with
  quadratic activation.
\newblock In {\em International Conference on Machine Learning}, 2018.

\bibitem{SimonDu19}
Simon~S. Du, Jason~D. Lee, Haochuan Li, Liwei Wang, and Xiyu Zhai.
\newblock Gradient descent finds global minima of deep neural networks.
\newblock In {\em International Conference on Machine Learning}, 2019.

\bibitem{Feizi}
Soheil Feizi, Hamid Javadi, Jesse Zhang, and David Tse.
\newblock Porcupine neural networks: (almost) all local optima are global.
\newblock {\em arXiv preprint arXiv:1710.02196}, 2017.

\bibitem{largscalelandscape}
Stanislav Fort and Stanislaw Jastrzebski.
\newblock Large scale structure of neural networks loss landscapes.
\newblock In {\em Advances in Neural Information Processing Systems}, 2019.

\bibitem{Topologygeometry}
C~Daniel Freeman and Joan Bruna.
\newblock Topology and geometry of half-rectified network optimization.
\newblock In {\em International Conference on Learning Representations}, 2017.

\bibitem{Gao}
Weihao Gao, Ashok~Vardhan Makkuva, Sewoong Oh, and Pramod Viswanath.
\newblock Learning one-hiddenlayer neural networks under general input
  distributions.
\newblock {\em arXiv preprint arXiv:1810.04133}, 2018.

\bibitem{modeconnectivity}
Timur Garipov, Pavel Izmailov, Dmitrii Podoprikhin, Dmitry~P Vetrov, and
  Andrew~G Wilson.
\newblock Loss surfaces, mode connectivity, and fast ensembling of dnns.
\newblock In {\em Advances in Neural Information Processing Systems}, 2018.

\bibitem{Matrixcompletion}
R.~Ge, J.~D. Lee, and T.~Ma.
\newblock Matrix completion has no spurious local minimum.
\newblock In {\em Advances in Neural Information Processing Systems}, 2016.

\bibitem{landdesign}
R.~Ge, J.~D. Lee, and T.~Ma.
\newblock Learning one-hidden-layer neural networks with landscape design.
\newblock {\em arXiv preprint arXiv:1711.00501}, 2017.

\bibitem{Goldblum2020}
M.~Goldblum, J.~Geiping, A.~Schwarzschild, M.~Moeller, and T.~Goldstein.
\newblock Truth or backpropaganda? an empirical investigation of deep learning
  theory.
\newblock In {\em International Conference on Learning Representations}, 2020.

\bibitem{Goodfellow}
Ian~J Goodfellow, Oriol Vinyals, and Andrew~M Saxe.
\newblock Qualitatively characterizing neural network optimization problems.
\newblock In {\em International Conference on Learning Representations}, 2015.

\bibitem{Hanin19}
Boris Hanin and David Rolnick.
\newblock Complexity of linear regions in deep networks.
\newblock In {\em International Conference on Machine Learning}, 2019.

\bibitem{Hanin19nips}
Boris Hanin and David Rolnick.
\newblock Deep relu networks have surprisingly few activation patterns.
\newblock In {\em Advances in Neural Information Processing Systems}, 2019.

\bibitem{HardtMa}
Moritz Hardt and Tengyu Ma.
\newblock Identity matters in deep learning.
\newblock In {\em International Conference on Learning Representations}, 2017.

\bibitem{Dacheng_Tao2020}
Fengxiang He, Bohan Wang, and Dacheng Tao.
\newblock Piecewise linear activations substantially shape the loss surfaces of
  neural networks.
\newblock In {\em International Conference on Learning Representations}, 2020.

\bibitem{NTK}
Arthur Jacot, Franck Gabriel, and Clement Hongler.
\newblock Neural tangent kernel: Convergence and generalization in neural
  networks.
\newblock In {\em Advances in Neural Information Processing Systems}, 2018.

\bibitem{Kawaguchi_nips16}
K.~Kawaguchi.
\newblock Deep learning without poor local minima.
\newblock In {\em Advances in Neural Information Processing Systems}, 2016.

\bibitem{KawaguchiBengio}
Kenji Kawaguchi and Yoshua Bengio.
\newblock Depth with nonlinearity creates no bad local minima in resnets.
\newblock {\em Neural Networks}, 118:167--174, 2019.

\bibitem{Kawaguchi19}
Kenji Kawaguchi and Leslie~Pack Kaelbling.
\newblock Elimination of all bad local minima in deep learning.
\newblock {\em arXiv preprint arXiv:1901.00279}, 2019.

\bibitem{DeepQuadratic}
Abbas Kazemipour, Brett Larsen, and Shaul Druckmann.
\newblock No spurious local minima in deep quadratic networks.
\newblock {\em arXiv preprint arXiv:2001.00098}, 2020.

\bibitem{AlexNet}
Alex Krizhevsky, Ilya Sutskever, and Geoffrey Hinton.
\newblock Imagenet classification with deep convolutional neural networks.
\newblock In {\em Advances in Neural Information Processing Systems}, 2012.

\bibitem{Laurent18}
Thomas Laurent and James~H. von Brecht.
\newblock Deep linear networks with arbitrary loss: All local minima are
  global.
\newblock In {\em International Conference on Machine Learning}, 2018.

\bibitem{Multilinear}
Thomas Laurent and James~H. von Brecht.
\newblock The multilinear structure of relu networks.
\newblock In {\em International Conference on Machine Learning}, 2018.

\bibitem{Li}
Dawei Li, Tian Ding, and Ruoyu Sun.
\newblock On the benefit of width for neural networks: Disappearance of bad
  basins.
\newblock {\em arXiv preprint arXiv:1812.11039}, 2018.

\bibitem{Visualizing}
Hao Li, Zheng Xu, Gavin Taylor, Christoph Studer, and Tom Goldstein.
\newblock Visualizing the loss landscape of neural nets.
\newblock In {\em Advances in Neural Information Processing Systems}, 2018.

\bibitem{Yuanzhi_Li_17}
Yuanzhi Li and Yang Yuan.
\newblock Convergence analysis of two-layer neural networks with relu
  activation.
\newblock In {\em Advances in Neural Information Processing Systems}, 2017.

\bibitem{AddingOneNeuron}
Shiyu Liang, Ruoyu Sun, Jason~D. Lee, and R.~Srikant.
\newblock Adding one neuron can eliminate all bad local minima.
\newblock In {\em Advances in Neural Information Processing Systems}, 2018.

\bibitem{LossSurfaceBinary}
Shiyu Liang, Ruoyu Sun, Yixuan Li, and R.~Srikant.
\newblock Understanding the loss surface of neural networks for binary
  classification.
\newblock In {\em International Conference on Machine Learning}, 2018.

\bibitem{Poggio}
Qianli Liao and Tomaso Poggio.
\newblock Theory of deep learning ii: Landscape of the empirical risk in deep
  learning.
\newblock {\em arXiv preprint arXiv:1703.09833}, 2017.

\bibitem{globallosslandscape_part1}
Bo~Liu.
\newblock Understanding global loss landscape of one-hidden-layer relu
  networks, part 1: theory.
\newblock {\em arXiv preprint arXiv:2002.04763}, 2020.

\bibitem{HaihaoLu}
Haihao Lu and Kenji Kawaguchi.
\newblock Depth creates no bad local minima.
\newblock {\em arXiv preprint arXiv:1702.08580}, 2017.

\bibitem{RINGINGRELUS2021}
Christian H.X.~Ali Mehmeti-Gopel, David Hartmann, and Michael Wand.
\newblock Ringing relus: Harmonic distortion analysis of nonlinear feedforward
  networks.
\newblock In {\em International Conference on Learning Representations}, 2021.

\bibitem{Mei}
S.~Mei, Y.~Bai, and A.~Montanari.
\newblock The landscape of empirical risk for non-convex losses.
\newblock {\em arXiv preprint arXiv:1607.06534}, 2016.

\bibitem{piecewiseconvex}
Tristan Milne.
\newblock Piecewise strong convexity of neural networks.
\newblock In {\em Advances in Neural Information Processing Systems}, 2019.

\bibitem{Montufar14}
Guido~F Montufar, Razvan Pascanu, Kyunghyun Cho, and Yoshua Bengio.
\newblock On the number of linear regions of deep neural networks.
\newblock In {\em Advances in Neural Information Processing Systems}, 2014.

\bibitem{Nguyen19}
Q.~Nguyen and M.~Hein.
\newblock On connected sublevel sets in deep learning.
\newblock In {\em International Conference on Machine Learning}, 2019.

\bibitem{Nguyenvalleys}
Quynh Nguyen, Mahesh~Chandra Mukkamala, and Matthias Hein.
\newblock On the loss landscape of a class of deep neural networks with no bad
  local valleys.
\newblock In {\em International Conference on Learning Representations}, 2019.

\bibitem{Nouiehed}
Maher Nouiehed and Meisam Razaviyayn.
\newblock Learning deep models: Critical points and local openness.
\newblock {\em arXiv preprint arXiv:1803.02968}, 2018.

\bibitem{RandomMatrix}
Jeffrey Pennington and Yasaman Bahri.
\newblock Geometry of neural network loss surfaces via random matrix theory.
\newblock In {\em International Conference on Machine Learning}, 2017.

\bibitem{Spectrum}
Jeffrey Pennington and Pratik Worah.
\newblock The spectrum of the fisher information matrix of a
  single-hidden-layer neural network.
\newblock In {\em Advances in Neural Information Processing Systems}, 2018.

\bibitem{SafranShamir16}
I.~Safran and O.~Shamir.
\newblock On the quality of the initial basin in overspecified neural networks.
\newblock In {\em International Conference on Machine Learning}, 2016.

\bibitem{SafranShamir18}
I.~Safran and O.~Shamir.
\newblock Spurious local minima are common in two-layer relu neural networks.
\newblock In {\em International Conference on Machine Learning}, 2018.

\bibitem{BoundingCounting}
Thiago Serra, Christian Tjandraatmadja, and Srikumar Ramalingam.
\newblock Bounding and counting linear regions of deep neural networks.
\newblock In {\em International Conference on Machine Learning}, 2018.

\bibitem{ResNetsProvablyBetter}
Ohad Shamir.
\newblock Are resnets provably better than linear predictors?
\newblock In {\em Advances in Neural Information Processing Systems}, 2018.

\bibitem{BoundsDescentPathsICLR20}
Arsalan Sharifnassab, Saber Salehkaleybar, and S.~Jamaloddin Golestani.
\newblock Bounds on over-parameterization for guaranteed existence of descent
  paths in shallow relu networks.
\newblock In {\em International Conference on Learning Representations}, 2020.

\bibitem{VGG}
Karen Simonyan and Andrew Zisserman.
\newblock Very deep convolutional networks for large-scale image recognition.
\newblock In {\em International Conference on Learning Representations}, 2015.

\bibitem{Soltanolkotabi}
M.~Soltanolkotabi, A.~Javanmard, and J.~D. Lee.
\newblock Theoretical insights into the optimization landscape of
  overparameterized shallow neural networks.
\newblock {\em IEEE Transactions on Information Theory}, 65(2):742--769, 2019.

\bibitem{SoudryCarmon}
D.~Soudry and Y.~Carmon.
\newblock No bad local minima: Data independent training error guarantees for
  multilayer neural networks.
\newblock {\em arXiv preprint arXiv:1605.08361}, 2016.

\bibitem{SoudryHoffer}
D.~Soudry and E.~Hoffer.
\newblock Exponentially vanishing suboptimal local minima in multilayer neural
  networks.
\newblock {\em arXiv preprint arXiv:1702.05777}, 2017.

\bibitem{GlobalLandscapeOverview}
Ruoyu Sun, Dawei Li, Shiyu Liang, Tian Ding, and R~Srikant.
\newblock The global landscape of neural networks: An overview.
\newblock {\em arXiv preprint arXiv:2007.01429}, 2020.

\bibitem{Swirszcz}
Grzegorz Swirszcz, Wojciech~Marian Czarnecki, and Razvan Pascanu.
\newblock Local minima in training of deep networks.
\newblock {\em arXiv preprint arXiv:1611.06310}, 2016.

\bibitem{CPWL}
J.~M. Tarela and M.~V. Martínez.
\newblock Region configurations for realizability of lattice piecewise-linear
  models.
\newblock {\em Mathematical and Computer Modeling}, 30:17–27, 1999.

\bibitem{Yuandong_Tian_17}
Yuandong Tian.
\newblock An analytical formula of population gradient for two-layered relu
  network and its applications in convergence and critical point analysis.
\newblock In {\em International Conference on Machine Learning}, 2017.

\bibitem{Venturi}
Luca Venturi, Afonso Bandeira, and Joan Bruna.
\newblock Spurious valleys in two-layer neural network optimization landscapes.
\newblock {\em arXiv preprint arXiv:1802.06384}, 2018.

\bibitem{linearpiecesCNN}
Huan Xiong, Lei Huang, Mengyang Yu, Li~Liu, Fan Zhu, and Ling Shao.
\newblock On the number of linear regions of convolutional neural networks.
\newblock In {\em International Conference on Machine Learning}, 2020.

\bibitem{GLOBALOPTIMALITY}
Chulhee Yun, Suvrit Sra, and Ali Jadbabaie.
\newblock Global optimality conditions for deep neural networks.
\newblock In {\em International Conference on Learning Representations}, 2018.

\bibitem{ChulheeYun19}
Chulhee Yun, Suvrit Sra, and Ali Jadbabaie.
\newblock Small nonlinearities in activation functions create bad local minima
  in neural networks.
\newblock In {\em International Conference on Learning Representations}, 2019.

\bibitem{Zhang}
Li~Zhang.
\newblock Depth creates no more spurious local minima.
\newblock {\em arXiv preprint arXiv:1901.09827}, 2019.

\bibitem{OnehiddenlayerReLUviaGD}
Xiao Zhang, Yaodong Yu, Lingxiao Wang, and Quanquan Gu.
\newblock Learning one-hidden-layer relu networks via gradient descent.
\newblock {\em arXiv preprint arXiv:1806.07808}, 2018.

\bibitem{Zhong17}
Kai Zhong, Zhao Song, Prateek Jain, Peter~L Bartlett, and Inderjit~S Dhillon.
\newblock Recovery guarantees for one-hidden-layer neural networks.
\newblock In {\em International Conference on Machine Learning}, 2017.

\bibitem{Zhou_Feng}
P.~Zhou and J.~Feng.
\newblock Empirical risk landscape analysis for understanding deep neural
  networks.
\newblock In {\em International Conference on Learning Representations}, 2018.

\bibitem{CRITICALPOINTS}
Yi~Zhou and Yingbin Liang.
\newblock Critical points of neural networks: Analytical forms and landscape
  properties.
\newblock In {\em International Conference on Learning Representations}, 2018.

\bibitem{Zhou_Liang_19}
Yi~Zhou, Junjie Yang, Huishuai Zhang, Yingbin Liang, and Vahid Tarokh.
\newblock Sgd converges to global minimum in deep learning via star-convex
  path.
\newblock In {\em International Conference on Learning Representations}, 2019.

\bibitem{QuanquanGu}
Difan Zou, Yuan Cao, Dongruo Zhou, and Quanquan Gu.
\newblock Stochastic gradient descent optimizes overparameterized deep relu
  networks.
\newblock {\em Machine Learning}, 109:467–492, 2020.

\end{thebibliography}

\end{document}